\documentclass[letterpaper, 10 pt, conference]{ieeeconf}  

\IEEEoverridecommandlockouts                              

\usepackage{graphicx}
\graphicspath{{figure/}}

\title{\LARGE \bf
Unified Structural–Hydrodynamic Modeling of Underwater Underactuated Mechanisms and Soft Robots}

\author{Chenrui Zhang$^{1\dagger}$, Graduate Student Member, IEEE, Yiyuan Zhang$^{1,2\dagger*}$, Graduate Student Member, IEEE,\\ Yunfei Ye$^{1}$, Junkai Chen$^{1}$, Haozhe Wang$^{1,2}$ and Cecilia Laschi$^{1}$, Fellow, IEEE
\thanks{$\dagger$These authors contributed equally to this work.}
\thanks{*Corresponding author: Yiyuan Zhang (yiyuan.zhang@u.nus.edu)}%
\thanks{$^{1}$Department of Mechanical Engineering and Advanced Robotics Centre, National University of Singapore, Singapore.}%
\thanks{$^{2}$Singapore-MIT Alliance for Research and Technology (SMART) Centre, 138602, Singapore.}%
\thanks{This work was supported by the Ministry of Education, Singapore, through the REBOT project (MOE-T2EP50221-0010); the Ministry of Foreign Affairs and International Cooperation, Italy, and the Agency for Science, Technology and Research, Singapore, through the DESTRO project (R22I0IR124); the National University of Singapore through the RoboLife project (WBS A-0009125-02-00, WBS A-0009125-03-00) and the LENUR project (WBS A-8004640-00-00), and the Bridging Fund project (WBS A-8003743-00-00); and by the National Research Foundation (NRF), Prime Minister’s Office, Singapore under its Campus for Research Excellence and Technological Enterprise (CREATE) programme. The Mens, Manus, and Machina (M3S) is an interdisciplinary research group (IRG) of the Singapore MIT Alliance for Research and Technology (SMART) centre.}
}
\usepackage{algorithm}
\usepackage{algorithmic}

\usepackage{booktabs}
\usepackage{amsmath}
\usepackage{amssymb}
\usepackage{amsfonts}
\usepackage{cite}
\begin{document}

\maketitle
\thispagestyle{empty}
\pagestyle{empty}

\begin{abstract}

Underwater robots are widely deployed for ocean exploration and manipulation. Underactuated mechanisms are particularly advantageous in aquatic environments, as reducing actuator count lowers the risk of motor leakage while introducing inherent mechanical compliance. However, accurate modeling of underwater underactuated and soft robotic systems remains challenging because it requires identifying a high-dimensional set of internal structural and external hydrodynamic parameters. In this work, we propose a trajectory-driven global optimization framework for unified structural–hydrodynamic modeling of underwater multibody systems. Inspired by the Covariance Matrix Adaptation Evolution Strategy (CMA-ES), the proposed approach simultaneously identifies coupled internal elastic, damping, and distributed hydrodynamic parameters through trajectory-level matching between simulation and experimental motion. This enables high-fidelity reproduction of both underactuated mechanisms and compliant soft robotic systems in underwater environments, using as little as a single video recording as input. We first validate the framework on a link-by-link underactuated multibody mechanism, demonstrating accurate identification of distributed hydrodynamic coefficients, with a normalized end-effector position error below 5\% across multiple trajectories, varying initial conditions, and both active–passive and fully passive configurations. The identified modeling strategy is further validated on an asymmetric octopus-inspired soft arm, confirming its effectiveness for compliant soft robotic systems in underwater environments. Finally, eight identified arms are assembled into a swimming octopus robot, where the unified parameter set enables realistic whole-body behavior without additional parameter retuning. These results demonstrate the scalability and transferability of the proposed structural–hydrodynamic modeling framework across underwater underactuated and soft robotic systems.

\end{abstract}

\section{INTRODUCTION}

Our world is mostly covered by water, which is abundant in resources and contains many fascinating unknowns \cite{ramirez-llodraDeepDiverseDefinitely2010}. Over the past decades, underwater robots have been increasingly adopted for underwater exploration to ensure the safety of human divers while enhancing human capability \cite{khatibOceanOneRobotic2016, mazzeoMarineRoboticsDeepSea2022}. Substantial knowledge has been accumulated in the design of underwater manipulators and vehicles, extending human capability in underwater manipulation, underwater locomotion, and observation \cite{morganAutonomousUnderwaterManipulation2022}. Underactuated mechanisms are particularly advantageous for underwater robotic systems, as they reduce the number of required actuators, thereby lowering the risk of motor leakage and improving structural compliance and robustness in complex aquatic environments \cite{mazzeoMarineRoboticsDeepSea2022}. Most aquatic animals, such as fish, sea lions, penguins, octopuses, and even worms, possess soft-bodied structures. At the same time, reducing control complexity and developing underactuated approaches. Bio-inspired soft robots derived from these organisms have therefore become a promising approach for developing improved underwater manipulators and swimming robots (vehicles) \cite{wangShallowWatersMariana2026, li_bioinspired_2026}. However, efficient modeling of the soft body itself is already a state-of-the-art research topic. Incorporating the hydrodynamics model of the surrounding fluid further increases the difficulty \cite{armaniniFlagellateUnderwaterRobotics2022, mathewReducedOrderModeling2025, laschiModelingEmbodiedIntelligence2023, liaoDynamicModelingPerformance2022, chenExplorationSwimmingPerformance2020}, where the hydrodynamics parameters typically relies on conventional computational fluid dynamics (CFD) methods \cite{zhongDesignModelingExperiment2024}, experimental method \cite{liuUnderwaterSoftArm2026} and further model-based identification \cite{armaniniFlagellateUnderwaterRobotics2022, liaoDynamicModelingPerformance2022, chenExplorationSwimmingPerformance2020}.

Inspired by the well-known efficient simplification pseudo-rigid approach \cite{armaniniSoftRobotsModeling2023} (the soft bodies are represented as series of rigid links, which are connected by joints) in soft robot modeling, researchers have begun developing robots in a similar discrete-body manner to facilitate direct modeling within the pseudo-rigid framework \cite{wangSpiRobsLogarithmicSpiralshaped2025, huang_physical_2026}. By leveraging the speed, accuracy, and computational efficiency of the MuJoCo multibody simulator \cite{todorovMuJoCoPhysicsEngine2012}, together with preset mechanical parameters, these models exhibit strong sim-to-real similarity in robotic manipulation. MuJoCo's relatively modest computational requirements further lower the barrier to simulation-based robotics research, making existing tools and knowledge more accessible across soft robotics and underactuated multibody systems.

\begin{figure*}[htbp]
    \centering
    \includegraphics[width=\textwidth]{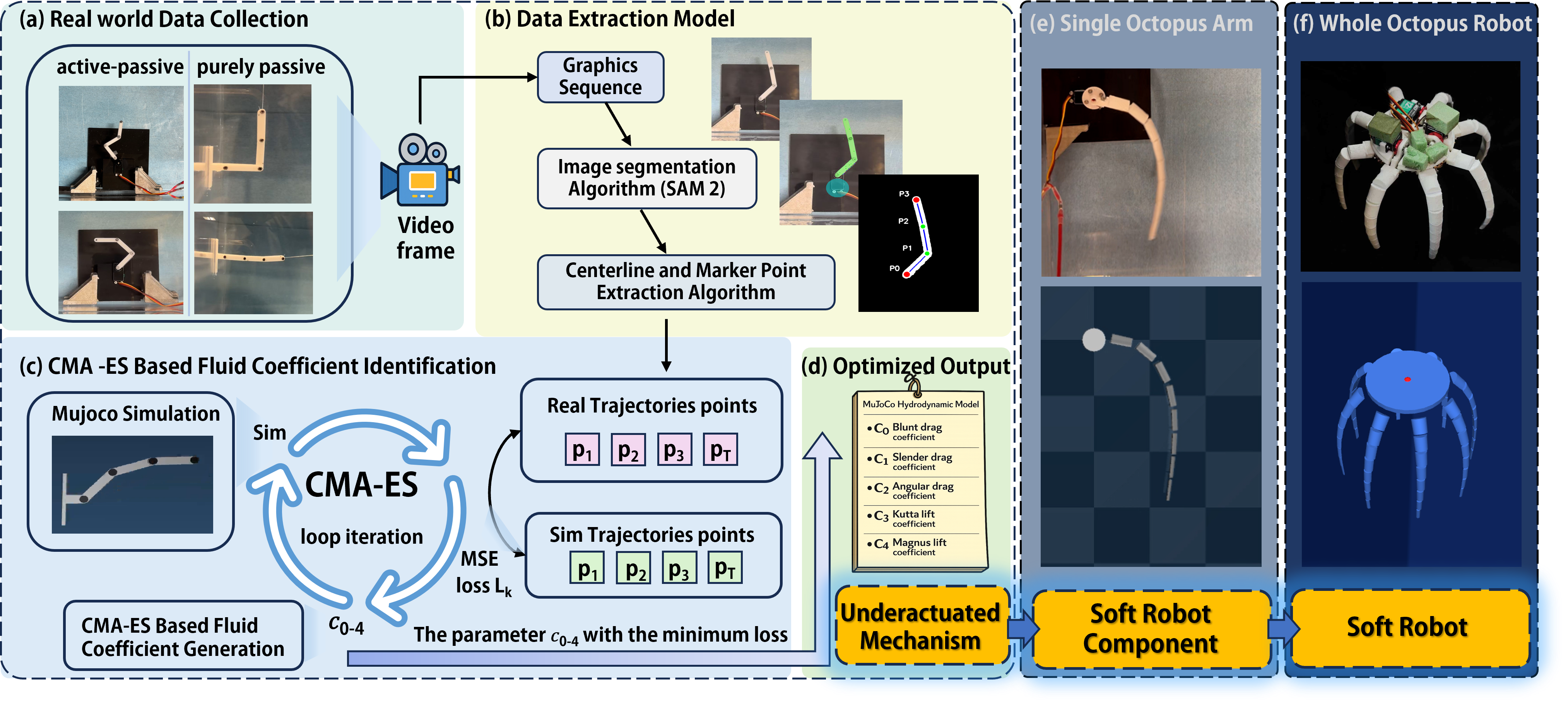}
    \caption{Trajectory-driven CMA-ES framework for simultaneous identification of distributed hydrodynamic coefficients, validated from a three-link mechanism to an octopus-inspired arm and an eight-arm robot. \textbf{(a) Real-world data collection:} Underwater trajectories are recorded using an active–passive coupled three-link mechanism and a purely passive mechanism. \textbf{(b) Data extraction:} SAM2-based segmentation and centerline extraction are used to obtain key trajectory points from experimental videos. \textbf{(c) Fluid coefficient identification:} Real trajectories are matched with MuJoCo simulations, and CMA-ES iteratively optimizes hydrodynamic parameters to minimize trajectory error. \textbf{(d) Optimized output:} The identified hydrodynamic coefficient set  $c_0$-$c_4$ calibrates the underwater dynamics model. \textbf{(e) Soft robot component:} The identified parameters are applied to a single octopus-inspired arm to evaluate parameter transferability. \textbf{(f) Soft robot:} Multiple identified arms are arranged in a circular configuration to construct an octopus-inspired robot, demonstrating scalability.}
    \label{fig:P1_whole}
\end{figure*}

In air, the current multibody simulation approaches can still be adjusted satisfactorily through manual tuning \cite{wang_openspirob} or a staged simulation-based parameter identification workflow through iterative calibration, where differential evolution is employed to optimize the bending stiffness and damping coefficients as internal soft-body parameters \cite{huang_physical_2026}. In water, however, the problem becomes significantly more challenging due to the introduction of hydrodynamics parameters, including blunt drag coefficient, slender drag coefficient, angular drag coefficient, kutta lift coefficient, and magnus lift coefficient. With such a high order of internal soft body dynamics parameters combined with external fluid environment dynamics parameters, especially in the underactuated and pseudo-rigid systems, manual or staged optimization approaches are no longer sufficient to characterize the physical dynamics parameters of underwater soft and underactuated robots.

Inspired by CMA-ES-based multibody dynamics parameter optimization for humanoid robots\cite{sobanbabu2025samplingbased}, we develop a trajectory-driven global optimization framework for identifying hydrodynamic parameters of underwater underactuated multibody systems. Starting from a passive three-link mechanism, we optimize MuJoCo’s ellipsoid-based fluid force coefficients, including drag, lift, and rotational fluid terms, to reproduce real-world motion trajectories with direction-dependent dynamics.

We then extend this identification approach to an octopus-inspired soft swimming arm modeled through a pseudo-rigid multibody representation. The optimized parameters enable accurate reproduction of experimental trajectories and consistent asymmetric dynamic responses. Finally, eight identical identified arms are arranged in a circular configuration to form an octopus-like swimming robot\cite{zhangOctopusSwimmingLikeRobotSoft2025a}, which shows strong visual agreement with the real robot without additional parameter fine-tuning.

Through this work, we aim to highlight our contribution to underwater robotics by empowering underwater underactuated system parameter identification with the latest global optimization approach, which enables simultaneous estimation of coupled parameters at the trajectory level. Furthermore, this work establishes hydrodynamics-based connection between underwater underactuated mechanisms and underwater soft robot pseudo-rigid modeling. By unifying internal structural dynamics and external fluid dynamic parameters within a single identification framework, we provide a systematic solution that bridges two traditionally separated modeling communities. We envision that our framework will provide an accessible and practical approach for both the underwater underactuated robotics community and the underwater soft robotics community.

\section{METHODS}

\subsection{Underactuated Mechanism Experimental Platforms}

This study focuses on the dynamic behaviors of underactuated underwater mechanisms. To systematically investigate active–passive underactuated coupling and purely passive hydrodynamic responses, two experimental platforms with distinct actuation configurations were developed. Both platforms were placed in a water tank, and their motions were recorded in real time using a high frame rate camera for subsequent data extraction and analysis.

The first platform, shown in Fig. 2(a), implements an active–passive coupled configuration. A motor directly actuates the first link segment, constituting the active component, while the subsequent links are not independently actuated and are mechanically connected to the driven segment. These downstream links interact with the surrounding fluid purely through passive dynamics induced by the upstream actuation. This configuration exhibits a high kinematic degree of freedom with a limited number of actuators, representing a canonical underactuated robotic structure. Therefore, it serves as a representative testbed for studying fluid–structure interactions in actively driven underactuated underwater robots.

To ensure safe and stable underwater operation, the motor and electronic components were waterproofed. Electrical connections were insulated and sealed using heat-shrink tubing to prevent water intrusion.

The second platform, shown in Fig. 2(b), represents a purely passive configuration. A supporting line initially holds the distal end of the uppermost link such that it maintains a perpendicular orientation relative to the second link. At the start of each experiment, the support line is released, allowing the multi-link structure to move freely under gravity without any active control input. The base frame is rigidly fixed to the sidewall of the water tank. This configuration enables evaluation of the intrinsic passive hydrodynamic response and provides baseline data for subsequent model validation and parameter identification.

\begin{figure}[htbp]
    \centering
    \includegraphics[width=\columnwidth]{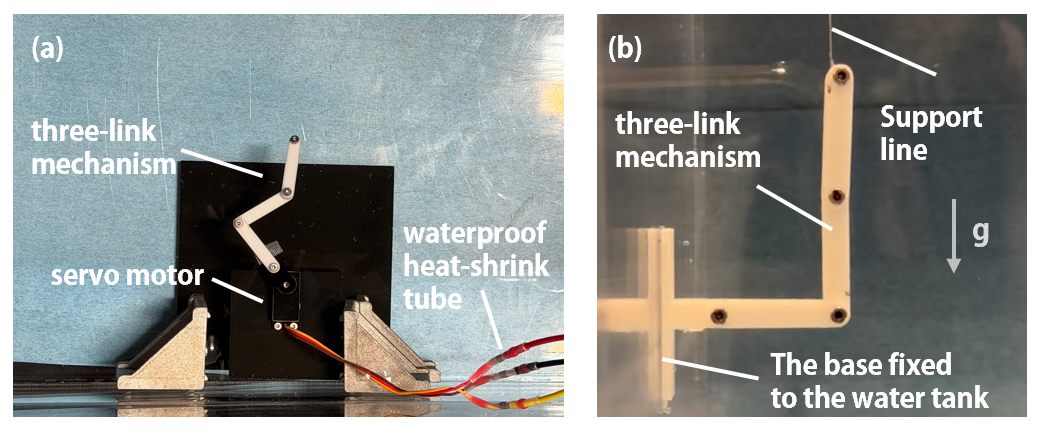}
    \caption{Three-link mechanism setup}
    \label{fig:P3}
\end{figure}

\subsection{Real-world Trajectory Extraction}

To obtain ground-truth motion trajectories (Fig. 3), we developed a vision-based trajectory extraction pipeline combining video segmentation and skeleton-based keypoint detection.

After each experiment, underwater motion was recorded using a fixed camera. The recorded video was converted into a sequence of image frames, where Fig. 3(a) illustrates a representative frame. The moving three-link structure was segmented using the SAM2 model \cite{ravi2024sam}, and the detailed procedure is described in Algorithm 1. As shown in Fig. 3(b), the dynamically captured link mask is highlighted in green, while the initial mask, indicated in blue, is introduced to exclude the motor screws from being incorrectly identified as marker points.

\begin{figure}[htbp]
    \centering
    \includegraphics[width=\columnwidth]{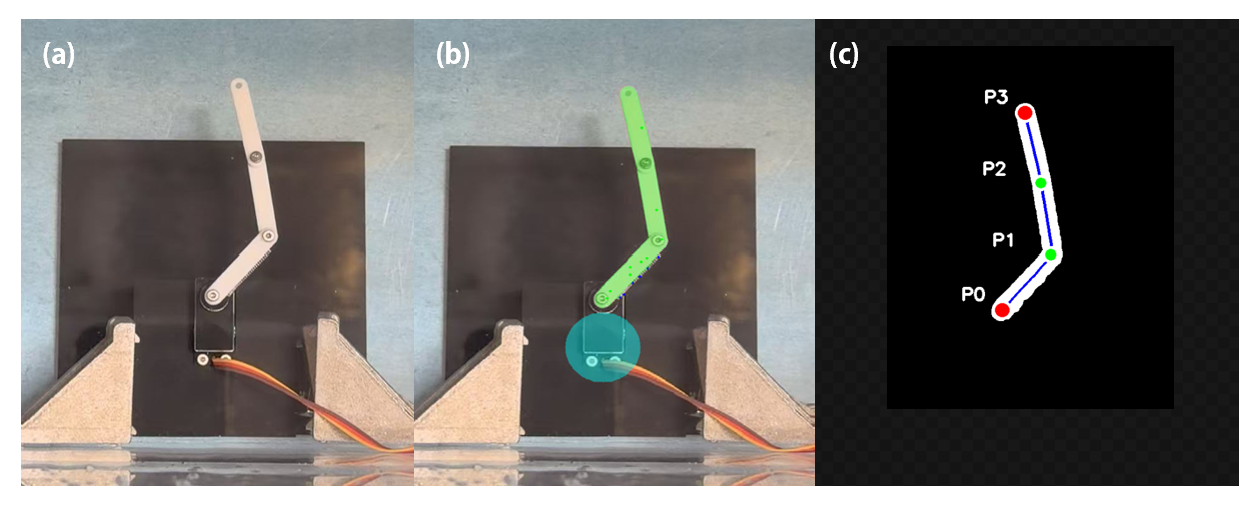}
    \caption{Vision-based trajectory extraction for the three-link mechanism}
    \label{fig:P2}
\end{figure}

Following segmentation, a skeletonization algorithm was applied to the binary mask to extract the medial axis of the link structure. Key trajectory points were then defined along the extracted centerline. For the motor-driven configuration, the lowest endpoint was defined as the reference $P_0$; for the gravity-driven passive configuration, the proximal endpoint was defined as $P_0$. The distal endpoint of the structure was defined as $P_3$. The centerline was subsequently divided into three equal segments to determine intermediate keypoints $P_1$ and $P_2$. In visualization, the two endpoints are marked in red, while the intermediate points are marked in green.

This processing pipeline was executed for all frames in the image sequence, resulting in time-series trajectory data that serve as ground truth for subsequent physical parameter identification and model validation.

\begin{algorithm}[htbp]
\caption{Trajectory Extraction via SAM2-based Video Segmentation}
\label{alg:sam2_traj}
\begin{algorithmic}[1]

\REQUIRE Image sequence $\{I_t\}_{t=0}^{T}$, static region mask $S$, user clicks $P$
\ENSURE Object trajectory $\mathcal{T} = \{p_t\}_{t=0}^{T}$

\STATE Initialize SAM2 video predictor $\mathcal{P}$
\STATE Select start frame $k$ and collect user annotations $P$

\STATE Initialize predictor with $(I_k, P)$ and obtain initial object mask $M_k$
\STATE Extract initial trajectory point $p_k \leftarrow \textsc{Centroid}(M_k \land \neg S)$

\FOR{$t = k+1$ \TO $T$}
    \STATE Propagate object mask $M_t \leftarrow \mathcal{P}(I_t \mid M_{t-1})$
    \STATE Remove static regions: $\tilde{M}_t \leftarrow M_t \land \neg S$
    \STATE Extract trajectory point $p_t \leftarrow \textsc{Centroid}(\tilde{M}_t)$
\ENDFOR

\STATE $\mathcal{T} \leftarrow \{p_k, p_{k+1}, \dots, p_T\}$
\RETURN $\mathcal{T}$

\end{algorithmic}
\end{algorithm}

\subsection{CMA-ES Based Fluid Coefficient Identification }
Hydrodynamic parameters of underwater multibody systems are difficult to obtain through direct measurement and cannot be manually assigned to simulation models in a physically consistent manner. Therefore, we employ MuJoCo's built-in ellipsoid-based fluid force model, which provides a unified approximation of distributed hydrodynamic effects for articulated bodies. Each link is parameterized by five fluid coefficients, including the blunt drag coefficient, slender drag coefficient, angular drag coefficient, Kutta lift coefficient, and Magnus lift coefficient. These coefficients are simultaneously identified through trajectory-level optimization.The physical parameter identification problem is formulated as a trajectory-matching optimization task. The objective is to recover hydrodynamic coefficients by minimizing the discrepancy between real-world and simulated trajectories.

\subsubsection{Parameter Modeling}

Consider an articulated underwater system composed of $n$ rigid links. Each link is associated with one observation point and characterized by five hydrodynamic coefficients. The complete parameter vector is defined as

\begin{equation}
\mathbf{x} =
\begin{bmatrix}
\mathbf{c}_1^\top &
\mathbf{c}_2^\top &
\cdots &
\mathbf{c}_n^\top
\end{bmatrix}^\top
\in \mathbb{R}^{5n},
\end{equation}

where $\mathbf{c}_i \in \mathbb{R}^{5}$ denotes the fluid coefficient vector of the $i$-th link. The optimization variable x is formed by concatenating the fluid coefficient vectors of all links into a single parameter vector. Given a parameter vector $\mathbf{x}$, the MuJoCo forward dynamics simulation produces a trajectory

\begin{equation}
\mathbf{T}_{\mathrm{sim}}(\mathbf{x})
=
\{ \mathbf{y}(t) \}_{t=1}^{N}
\in \mathbb{R}^{N \times 3n},
\end{equation}

where $\mathbf{y}(t)$ stacks the positions of all observation points at time step $t$. The corresponding real-world trajectory is denoted as $\mathbf{T}_{\text{real}}$.

\subsubsection{Loss Function and Optimization Objective}

The discrepancy between simulated and real trajectories is quantified using the mean squared error (MSE):

\begin{equation}
L(\mathbf{x})
=
\frac{1}{N}
\sum_{t=1}^{N}
\left\|
\mathbf{y}_{\mathrm{sim}}(t;\mathbf{x})
-
\mathbf{y}_{\mathrm{real}}(t)
\right\|^2.
\end{equation}

The parameter identification problem is therefore formulated as

\begin{equation}
\mathbf{x}^*
=
\arg\min_{\mathbf{x}} L(\mathbf{x}),
\quad
\text{s.t.} \quad
\mathbf{b}_{\min} \le \mathbf{x} \le \mathbf{b}_{\max},
\end{equation}

where $\mathbf{b}_{\min}$ and $\mathbf{b}_{\max}$ define physically feasible bounds.

\subsubsection{CMA-ES-Based Optimization}

Due to the highly nonlinear and non-differentiable nature of fluid–structure interactions in underwater dynamics, gradient-based methods are not suitable. Therefore, we employ the CMA-ES for derivative-free global optimization.

At generation $g$, candidate solutions are sampled from a multivariate normal distribution:

\begin{equation}
\mathbf{x}_k^{(g)}
\sim
\mathcal{N}
\left(
\mathbf{m}^{(g)},
\left(\sigma^{(g)}\right)^2 \mathbf{C}^{(g)}
\right),
\quad
k = 1, \dots, \lambda.
\end{equation}

For each sampled parameter vector, the following steps are performed:

\begin{itemize}
    \item The parameters are loaded into the MuJoCo model;
    \item Forward dynamics simulation is executed;
    \item The trajectory loss $L(\mathbf{x})$ is computed;
    \item The sampling distribution is updated according to fitness ranking.
\end{itemize}

\begin{algorithm}[htbp]
\caption{CMA-ES Based Fluid Coefficient Identification in MuJoCo}
\label{alg:cmaes_identification}
\begin{algorithmic}[1]

\STATE \textbf{Input:}
\STATE \quad Target trajectory $\mathbf{T}_{\text{real}}$ from real experiments
\STATE \quad Initial parameter guess $\mathbf{x}_0$
\STATE \quad Parameter bounds $[\mathbf{b}_{\min}, \mathbf{b}_{\max}]$
\STATE \quad Maximum evaluations $N_{\max}$

\STATE \textbf{Output:}
\STATE \quad Optimized fluid coefficients $\mathbf{x}^*$

\STATE Load real trajectory data and construct target trajectory matrix $\mathbf{T}_{\text{real}}$
\STATE Initialize CMA-ES with dimension $D=15$, step size $\sigma_0$, and population size $\lambda$
\STATE Initialize evaluation cache $\mathcal{C}$

\WHILE{not converged \textbf{and} $n < N_{\max}$}
    \STATE Generate candidate solutions $\{\mathbf{x}_k\}$
    \FOR{each candidate $\mathbf{x}_k$}
        \STATE Enforce physical and boundary constraints on $\mathbf{x}_k$
        \STATE Update MuJoCo XML fluid coefficients
        \STATE Run simulation and sample trajectory $\mathbf{T}_{\text{sim}}$
        \STATE Compute MSE loss $L_k$
        \STATE Cache evaluation results
    \ENDFOR
    \STATE Update CMA-ES distribution
\ENDWHILE

\STATE \textbf{return} best parameters $\mathbf{x}^*$

\end{algorithmic}
\end{algorithm}

\subsection{Optimized Output}

The optimization is performed for up to 5000 iterations, yielding the parameter set $\mathbf{x}^*$ that minimizes the trajectory discrepancy. The overall real-to-sim identification framework is illustrated in Fig.~1(c), and implementation details are provided in Algorithm 2.

This approach establishes a closed-loop trajectory-driven simulation calibration framework, forming the foundation of the proposed real-to-sim system.

After iterative optimization, CMA-ES converges to an optimal parameter set 
$X^*$ within the bounded search space, yielding identified hydrodynamic coefficients $c_0$-$c_4$ for each link. This parameter set minimizes the mean squared trajectory discrepancy over the entire time horizon.

By incorporating the identified coefficients into the MuJoCo model, we obtain the final calibrated underwater simulation model. The resulting model reproduces the experimentally observed motion patterns and dynamic responses without further manual tuning, serving as the baseline simulation model for subsequent analysis and real-to-sim validation.

\section{Real-to-Sim Underactuated Mechanism Experiments}

To evaluate the generalization capability of the identified hydrodynamic parameters and the fidelity of the calibrated simulation model, we conducted four independent experiments across two system configurations: the active–passive coupled mechanism and the purely passive mechanism. For each configuration, two distinct initial poses were tested. The results are presented in Figs. 4–7.

For each experimental group: Fig. (a) shows the initial pose of the real system; Fig. (b) shows the corresponding initial pose in simulation with the identified parameters; Fig. (c) illustrates the overall link motion in the real experiment, generated by superimposing frame-wise trajectory points extracted; Figs. (d), (e), and (f) present the time-series trajectories of keypoints $P_1$ and $P_2$, and $P_3$, respectively. In the trajectory comparison plots, the blue solid lines denote real-world measurements, while the red dashed lines represent simulation results using the identified parameter set.

\begin{figure}[htbp]
    \centering
    \includegraphics[width=\columnwidth]{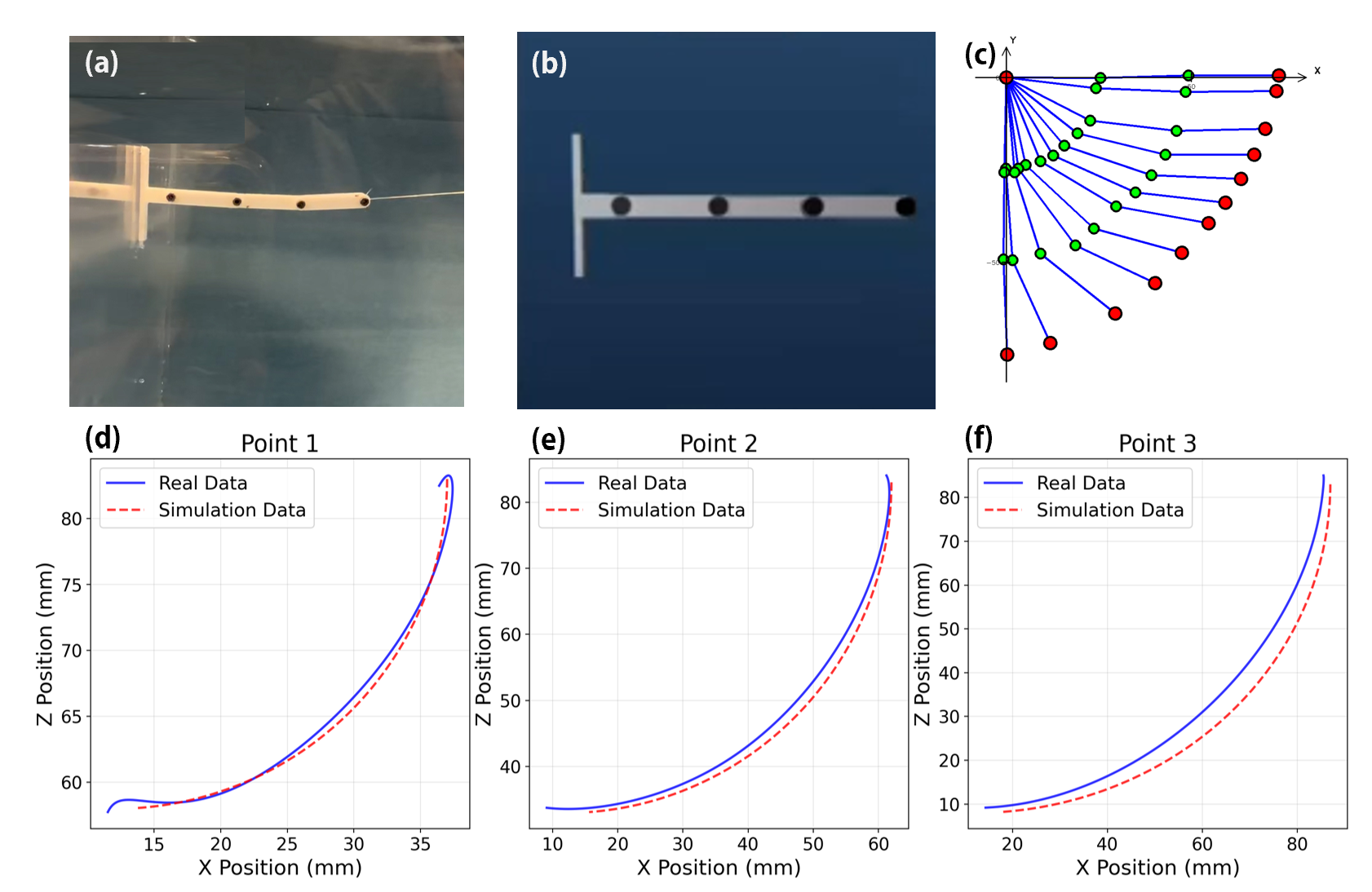}
    \caption{Horizontal three-link trajectories: extracted vs. real data}
    \label{fig:P42}
\end{figure}
\begin{figure}[htbp]
    \centering
    \includegraphics[width=\columnwidth]{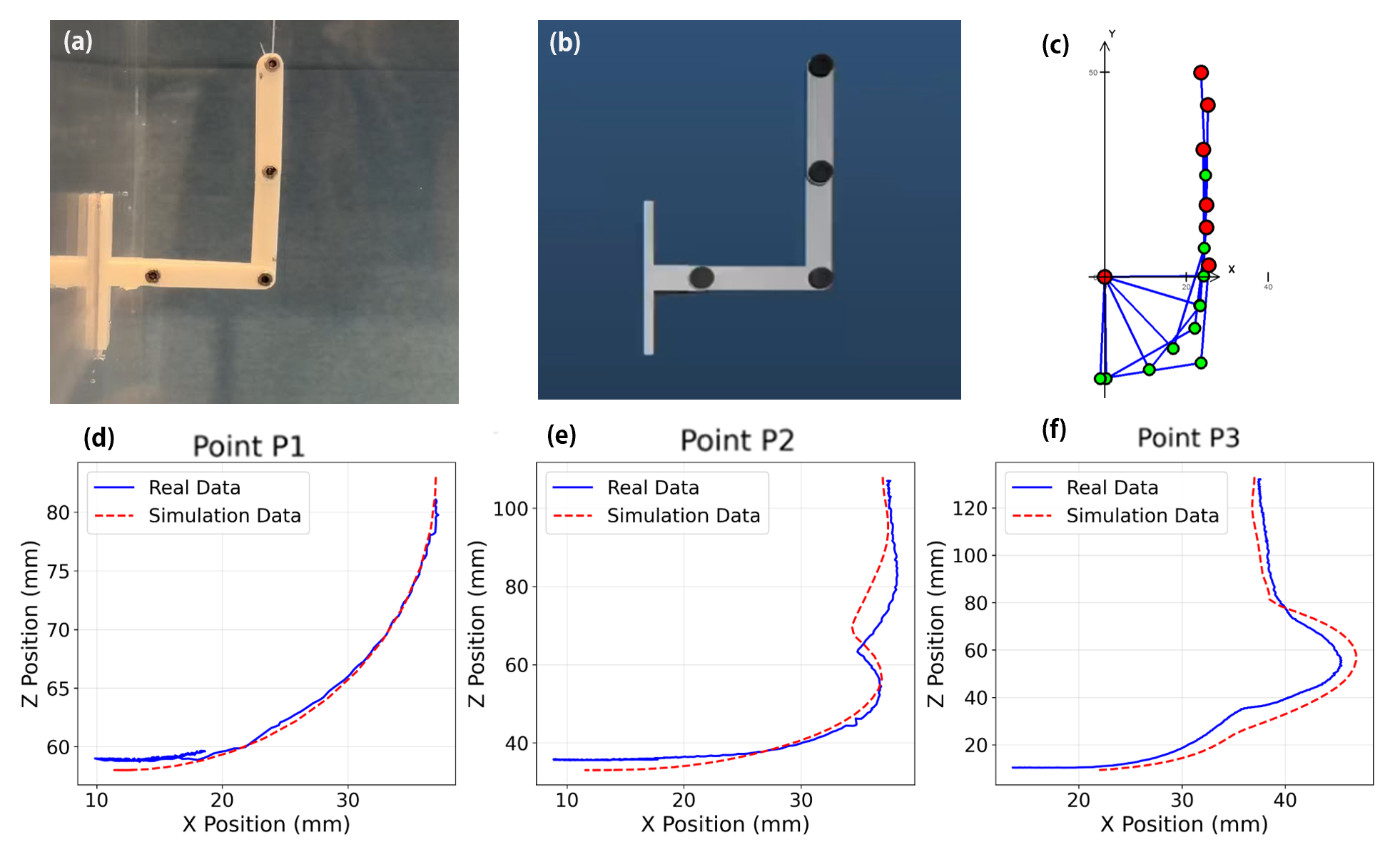}
    \caption{Right-angle three-link trajectories: extracted vs. real data}
    \label{fig:P52}
\end{figure}
\begin{figure}[htbp]
    \centering
    \includegraphics[width=\columnwidth]{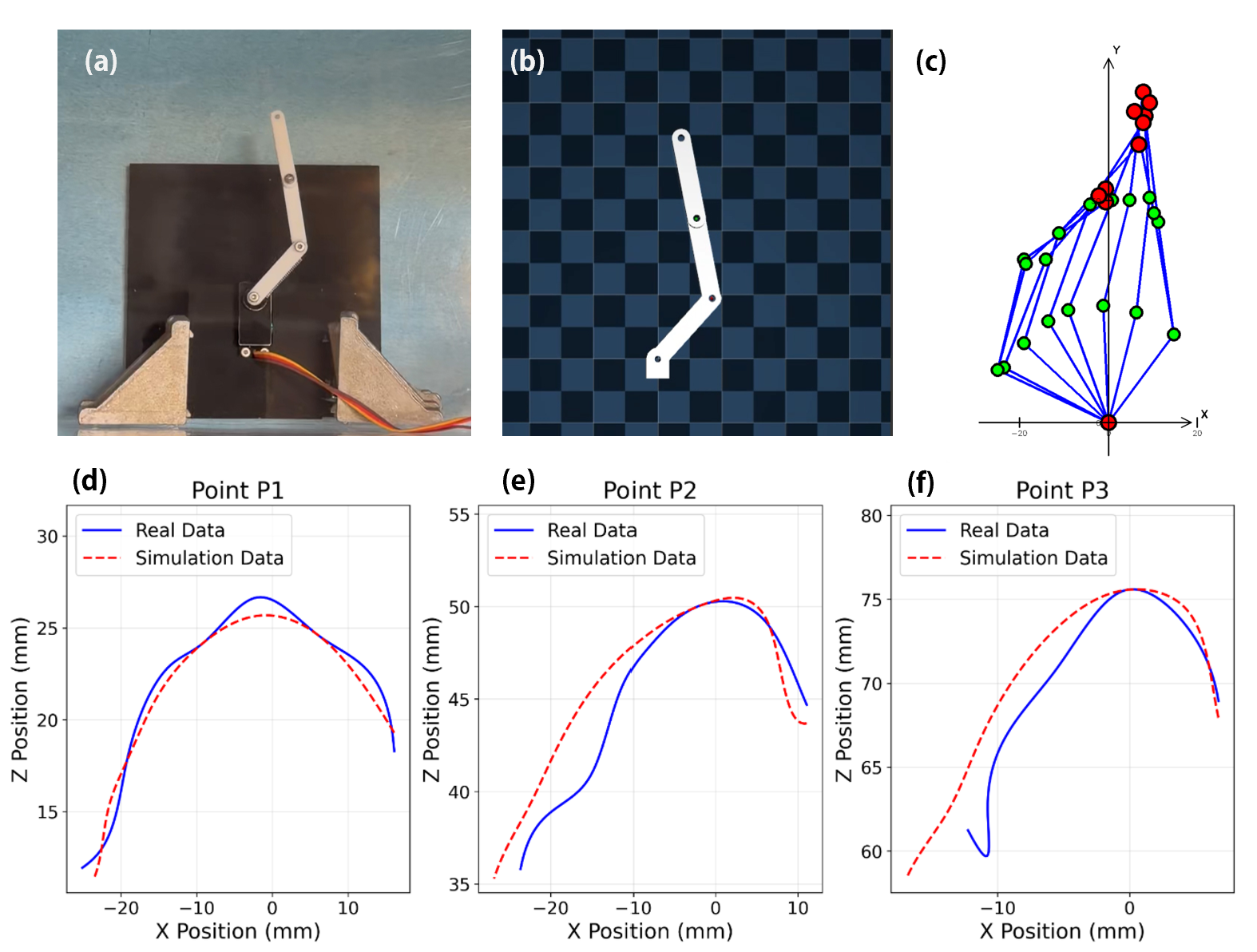}
    \caption{Motor-driven three-link pose I: extracted trajectories vs. real data}
    \label{fig:P62}
\end{figure}
\begin{figure}[htbp]
    \centering
    \includegraphics[width=\columnwidth]{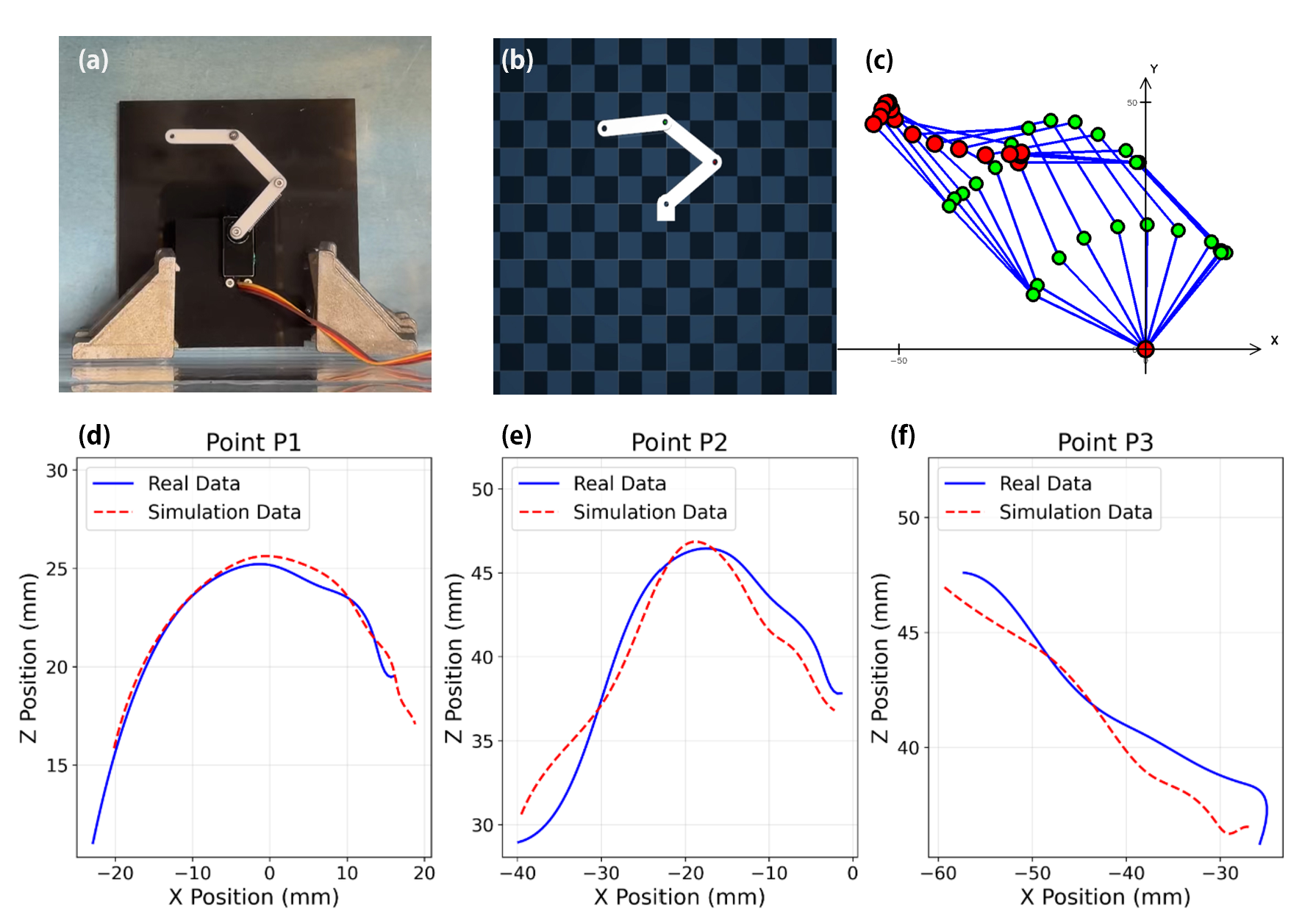}
    \caption{Motor-driven three-link pose II: extracted trajectories vs. real data}
    \label{fig:P72}
\end{figure}

In the real-to-sim calibration process, the MuJoCo model incorporates both directly specified physical parameters and parameters identified through the proposed optimization framework.

Several parameters are assigned based on known physical properties or standard simulation settings. For the underwater environment, the fluid viscosity is directly specified according to the experimental conditions. For the actuation system, motor-related parameters including the velocity and the force range are directly adopted from the actuator specifications. For the rigid-link structures, collision-related properties  are enabled to allow geometric collision detection within the simulation environment. 

The key parameters identified by the proposed method are the hydrodynamic coefficients governing the interaction between the links and the surrounding fluid. Specifically, five fluid coefficients $(c_0$–$c_4)$ for each link are optimized using the CMA-ES based identification framework described in Section II-C. These parameters characterize the dominant hydrodynamic forces acting on each link and are estimated by minimizing the trajectory discrepancy between the simulated and real-world motion data. In addition, joint damping and friction loss parameters are also identified to ensure stable numerical integration and realistic mechanical behavior.

By integrating the mechanical parameters with the identified hydrodynamic coefficients, the resulting MuJoCo model forms the final calibrated simulation system used for the subsequent real-to-sim validation experiments.

Across all four experimental scenarios, the calibrated simulation model consistently reproduces the motion trends and spatial evolution observed in the real system under varying initial conditions. In the active–passive configuration, the amplitude modulation of the distal keypoint $P_2$ and $P_3$ closely match experimental observations. In the purely passive configuration, the gravity-driven oscillatory decay behavior is likewise captured by the simulation model.

These results demonstrate that the proposed trajectory-based parameter identification method achieves not only trajectory fitting for a single experiment but also robust real-to-sim consistency across varying initial conditions, indicating strong physical validity and generalization capability of the identified parameters.

Table I summarizes the average trajectory errors across the four experimental configurations. The error metric is defined as the Euclidean distance between simulated and real keypoint positions under time-aligned conditions, averaged over the entire trajectory duration.

\begin{table}[htbp]
\centering
\caption{Underactuated Robot Component Results}
\label{tab:underactuated_component}
\begin{tabular}{lccc}
\toprule
Case & P1 & P2 & P3 \\
\midrule
Purely Passive Case 1 & 0.87 & 1.53 & 2.79 \\
Purely Passive Case 2 & 1.35 & 1.76 & 2.54 \\
Active-Passive Case 1 & 1.62 & 2.78 & 2.55 \\
Active-Passive Case 2 & 1.14 & 1.97 & 1.68 \\
\bottomrule
\end{tabular}
\end{table}
In the experimental mechanism, each link has a length of 30 mm, with an overlap radius of 5 mm between adjacent links, resulting in an overall effective length of approximately 80 mm. Normalizing the trajectory error relative to this characteristic system length provides a physically meaningful performance indicator.

Across all four initial conditions, the overall average error remains below 5\% of the total system length. Considering the strong nonlinear fluid–structure coupling and high-dimensional parameter space inherent to underwater underactuated mechanisms, this error level indicates that the identified hydrodynamic parameters accurately capture the real dynamic behavior.

These results validate the effectiveness and robustness of the proposed trajectory-based parameter identification method for underwater underactuated mechanisms and establish a reliable modeling foundation for extending the approach to fully actuated underwater robotic systems.

\section{Soft Robot Component Experiment}

In the previous sections, the proposed physical parameter identification method demonstrated strong performance on an underwater underactuated three-link mechanism. In this section, the method is further extended to an actual underwater soft robotic component to evaluate its applicability and generalization capability in complex flexible structures and interactive fluid dynamics.

Under the same active–passive actuation strategy, the original three-link mechanism is replaced by a single arm of an octopus-inspired robot (see Fig. 8(a)). The arm consists of eight serially connected segments with gradually decreasing width along its length, forming a tapered morphology. Compared with the rigid three-link mechanism, this structure exhibits higher degrees of freedom and stronger nonlinear fluid–structure coupling effects, significantly increasing the difficulty of physical parameter identification.

\begin{figure}[htbp]
    \centering
    \includegraphics[width=\columnwidth]{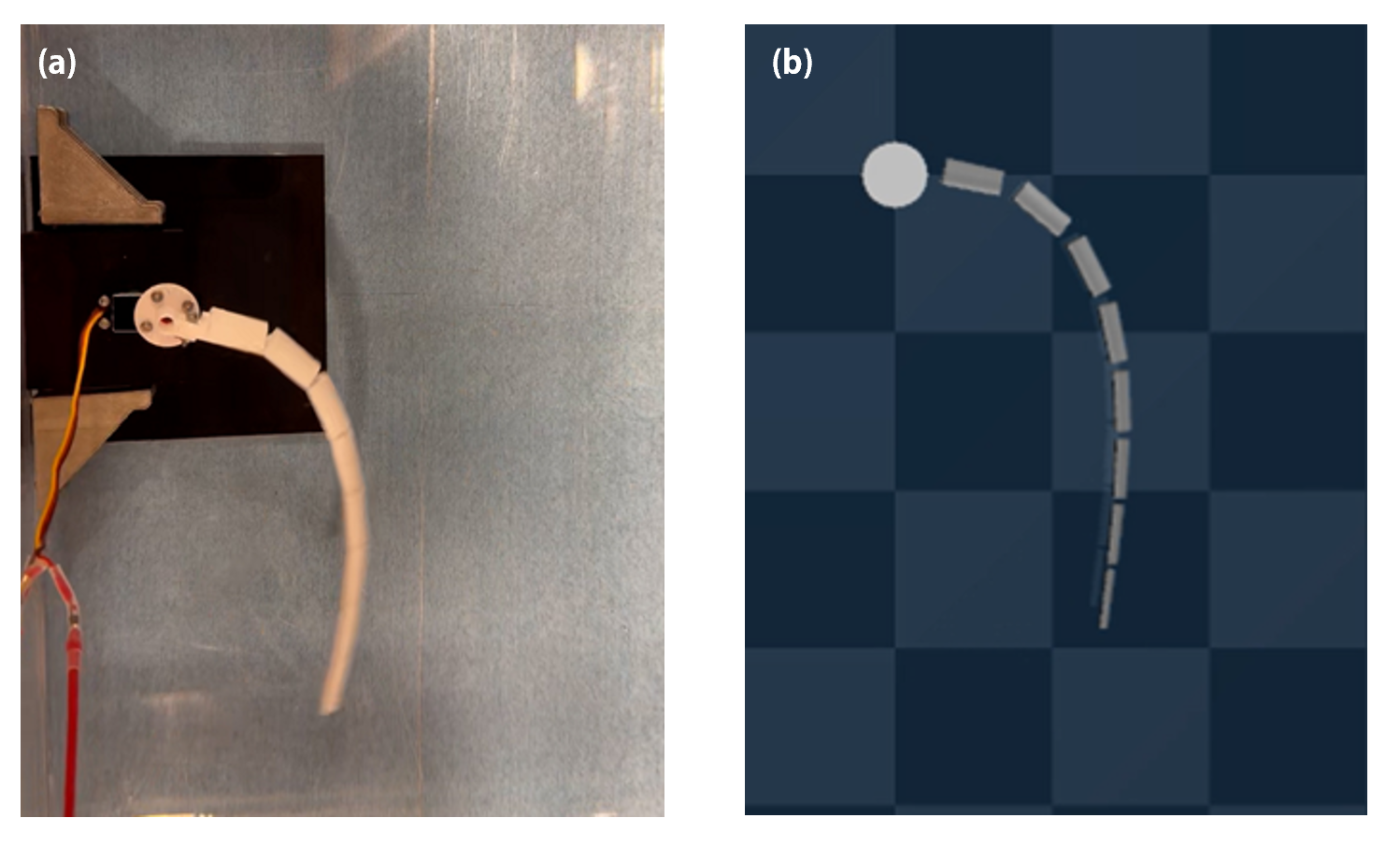}
    \caption{Soft robot component prototype and its MuJoCo simulation model}
    \label{fig:P8}
\end{figure}
During underwater locomotion, the octopus-inspired arm demonstrates clear direction-dependent motion characteristics. When the direction of motion changes, both the velocity and posture differ, while the overall motion maintains a 2:1 velocity ratio. Specifically, the duration of the servo-driven counterclockwise motion is twice that of the clockwise motion. Fig. 9(a) and Fig. 9(b) illustrate the real experimental results. It can be observed that during counterclockwise motion, the arm undergoes significant bending deformation, whereas during clockwise motion, the arm remains relatively straight while swinging back.
\begin{figure}[htbp]
    \centering
    \includegraphics[width=0.9\columnwidth]{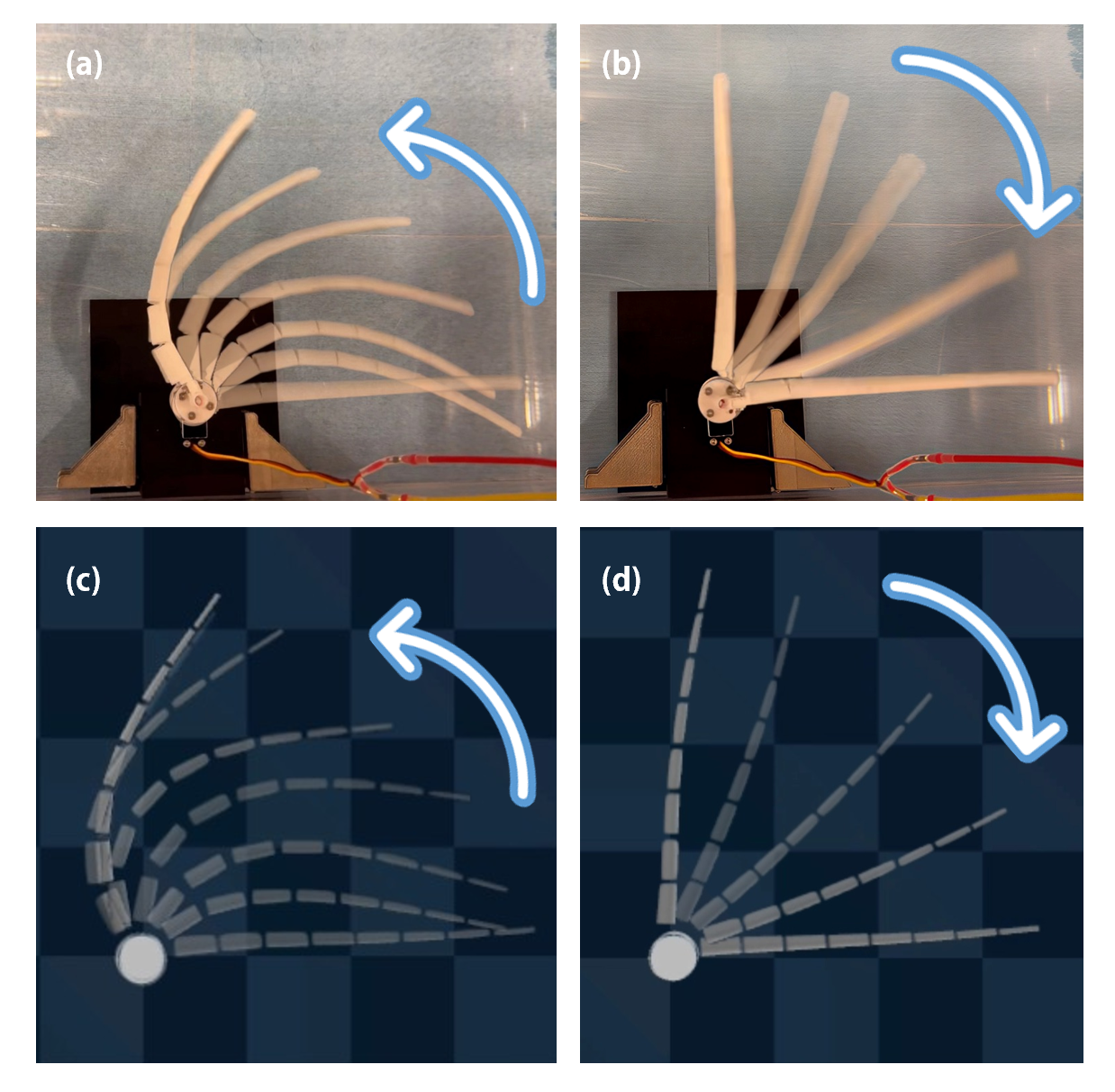}
    \caption{Soft robot component motion: prototype vs. MuJoCo simulation}
    \label{fig:P9}
\end{figure}

Using the previously identified underwater fluid dynamic parameters, simulation experiments were conducted under the same temporal conditions as the real system. Fig. 9(c) and Fig. 9(d), together with the supplementary experimental video, present the simulation results. The simulation accurately captures the direction-dependent postural variations. Under identical time conditions, the simulated and real motions exhibit high consistency, with clear one-to-one correspondence throughout the motion process.

Fig. 10(a) and Fig. 10(b) adopt the previously described Real-world Trajectory Extraction method, where the distal endpoint of the fourth segment is defined as $P_1$ , and the tip of the entire arm is defined as $P_2$. These results (Fig. 10(c) and Fig. 10(d)) indicate that the proposed physical parameter identification method is not limited to underactuated rigid mechanisms but can also effectively model the dynamic behavior of complex flexible driven components in underwater environments, thereby providing a foundation for its application in underwater robotic systems.
\begin{figure}[htbp]
    \centering
    \includegraphics[width=\columnwidth]{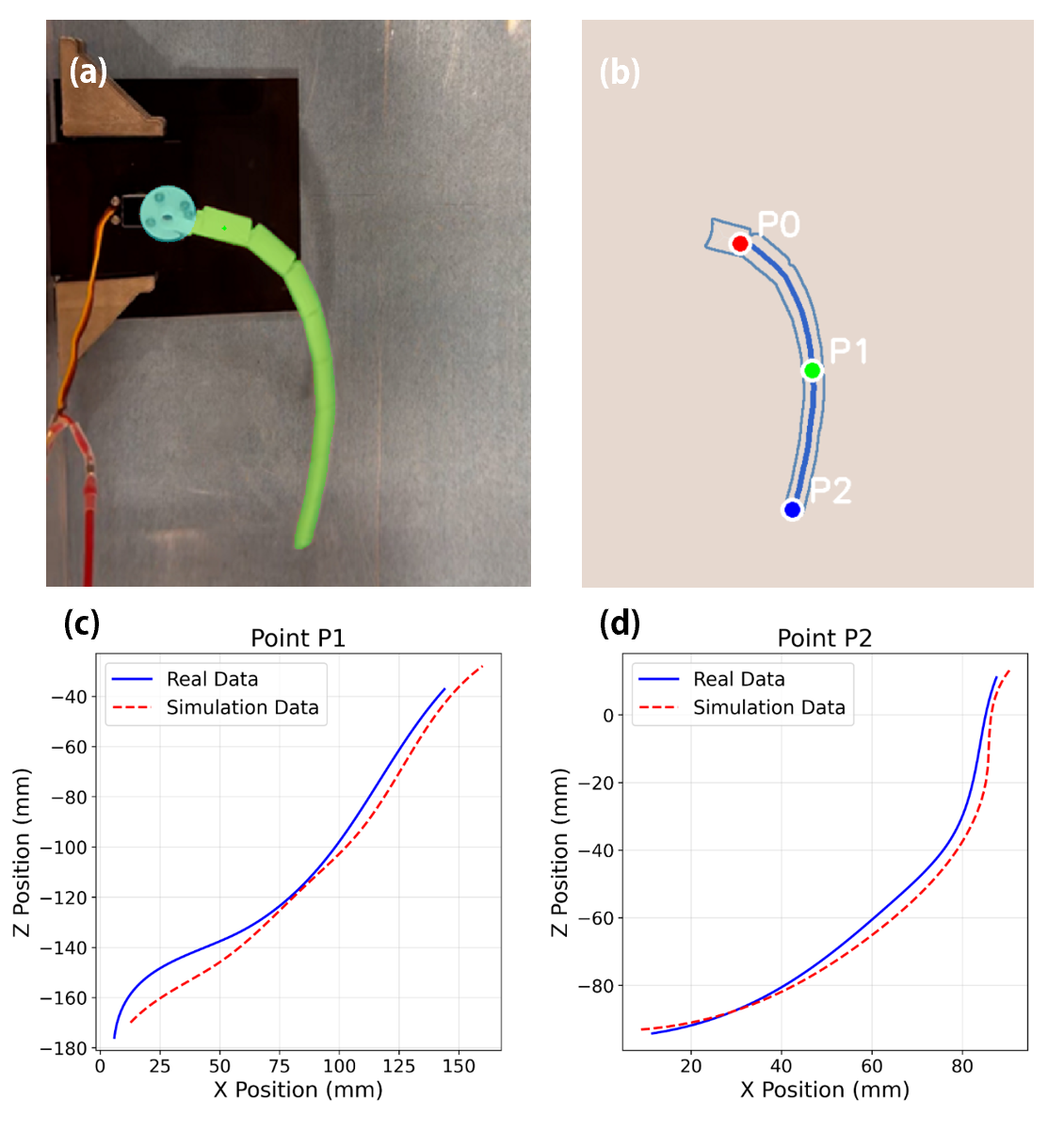}
    \caption{Soft robot component: extracted trajectories vs. real data }
    \label{fig:P12}
\end{figure}

\section{Soft Robot Demonstration}

After achieving satisfactory validation results on the single robotic arm, the proposed physical parameter identification method was further extended to the full robot body. Based on structural designs reported in the literature, a complete octopus-inspired underwater robot was constructed (see Fig. 11(a)), consisting of a central body platform and eight identical soft arms.

\begin{figure}[htbp]
    \centering
    \includegraphics[width=\columnwidth]{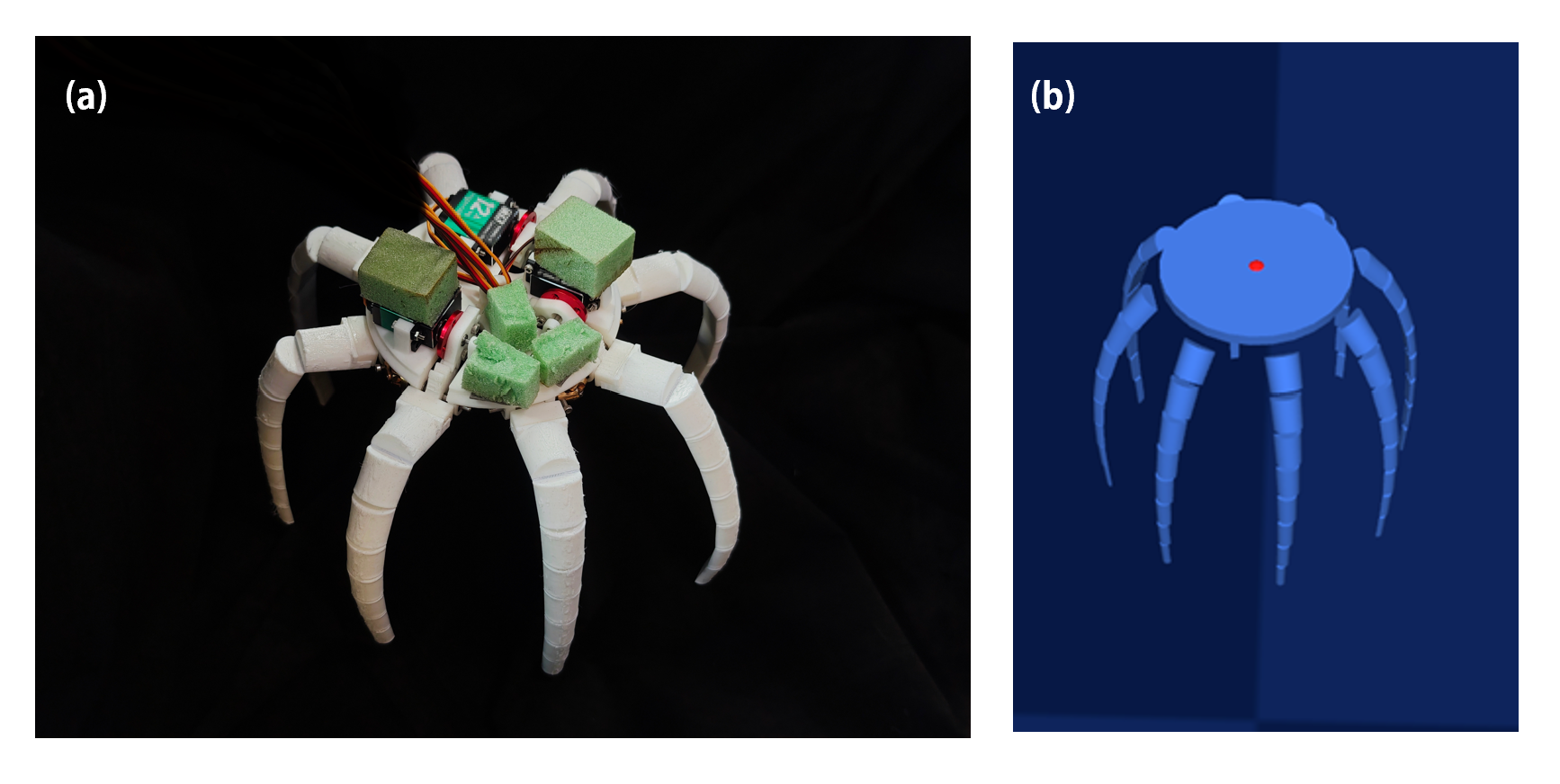}
    \caption{Octopus-inspired soft robot prototype and its MuJoCo simulation model}
    \label{fig:P10}
\end{figure}

In the full-body simulation model, the previously identified fluid dynamic parameters of a single arm were directly applied to all eight arms. A synchronized actuation strategy was adopted to drive the arms simultaneously. Through comparative experiments between the real robot and the simulation model (see Fig. 12), it can be observed that the simulated robot exhibits motion patterns, postural evolution, and propulsion characteristics highly consistent with those of the physical robot. This demonstrates that the identified parameters maintain good generalization capability and stability in a multi-actuator coupled system.

It should be noted that, under identical temporal conditions, the simulated robot achieves a slightly shorter displacement than the real robot. This discrepancy is primarily attributed to the fact that only the arm structures were subjected to physical parameter identification, while the central body platform was not calibrated for fluid dynamic parameters. Since the platform also contributes additional mass and hydrodynamic drag effects during underwater locomotion, the unmodeled dynamics accumulate over time, resulting in reduced simulated displacement.

Despite this difference, the overall motion trends and postural evolution remain highly consistent, further validating the scalability and engineering applicability of the proposed physical parameter identification method to complete underwater robotic systems.

\begin{figure*}[htbp]
    \centering
    \includegraphics[width=\textwidth]{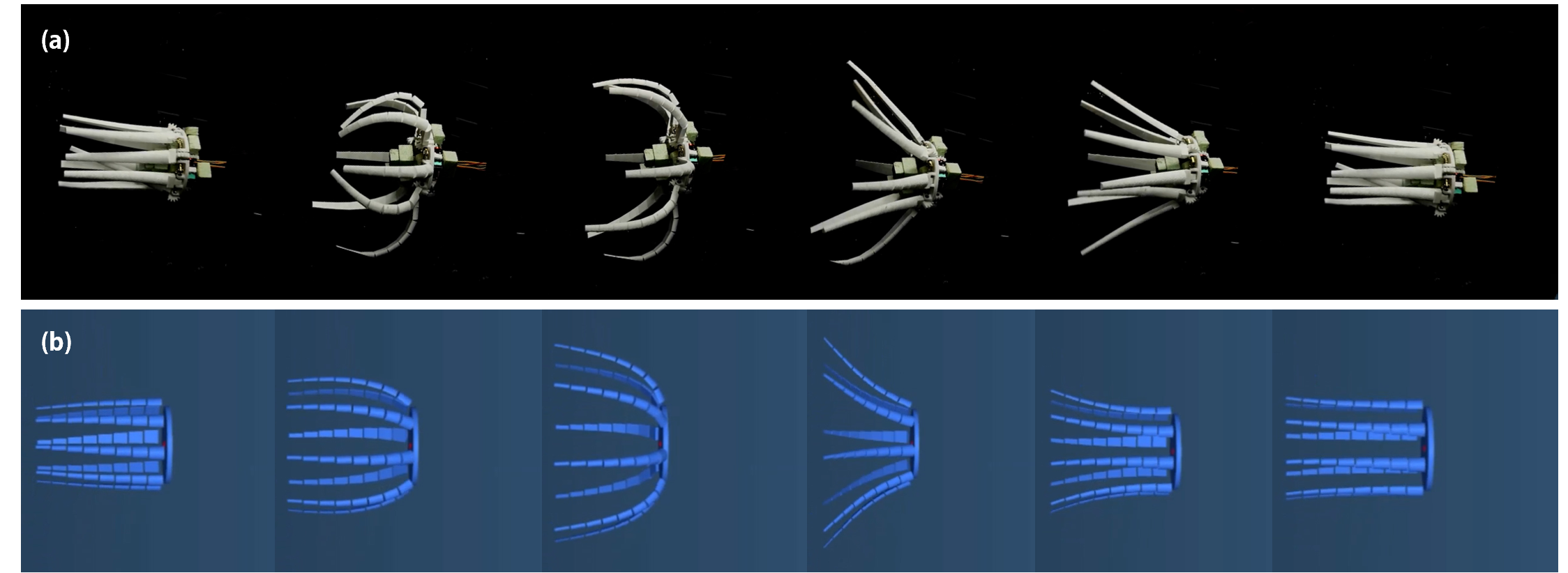}
    \caption{Octopus-inspired soft robot motion: prototype vs. MuJoCo simulation}
    \label{fig:P122_whole}
\end{figure*}
\section{CONCLUSIONS}

In this work, we propose a unified physical parameter identification method for underwater underactuated systems, enabling consistent real-to-sim transfer from simplified mechanisms to complete robotic platforms. Unlike conventional sequential identification strategies, the proposed approach simultaneously estimates multiple fluid dynamic parameters within a single framework, thereby avoiding error accumulation and parameter coupling distortions introduced by stepwise calibration.

Overall, the results demonstrate strong real-to-sim consistency across simplified mechanisms, soft robot components, and complete robotic platforms. More broadly, this work establishes a practical workflow for soft robotic simulation in fluid environments, spanning real-world trajectory collection, trajectory-driven hydrodynamic parameter identification in MuJoCo, and transfer of the identified parameters from simplified mechanisms to soft robot components and full robotic platforms. By integrating these steps into a unified framework, we provide a feasible route toward realistic fluid-environment simulation of soft robots, while MuJoCo’s computational efficiency and low resource requirements further support broader adoption.

Future work will focus on extending the proposed framework to global hydrodynamic parameter identification of the complete robot structure under flowing-water conditions, improving real-to-sim fidelity for underwater robotic systems.

\addtolength{\textheight}{-0cm}   


\section{ACKNOWLEDGMENT}

The authors acknowledge the use of AI-assisted tools, including ChatGPT and Cursor, during the preparation of this manuscript. These tools were utilized to support tasks such as language refinement, figure preparation (Fig.1. Fig.4-7), and assistance in experimental code development. All technical content, experimental design, results analysis, and conclusions were developed and verified by the authors.


\bibliographystyle{IEEEtran}
\bibliography{references}

@article{armaniniFlagellateUnderwaterRobotics2022,
  title = {Flagellate {{Underwater Robotics}} at {{Macroscale}}: {{Design}}, {{Modeling}}, and {{Characterization}}},
  shorttitle = {Flagellate {{Underwater Robotics}} at {{Macroscale}}},
  author = {Armanini, Costanza and Farman, Madiha and Calisti, Marcello and {Giorgio-Serchi}, Francesco and Stefanini, Cesare and Renda, Federico},
  year = 2022,
  month = apr,
  journal = {IEEE Transactions on Robotics},
  volume = {38},
  number = {2},
  pages = {731--747},
  issn = {1941-0468},
  doi = {10.1109/TRO.2021.3094051}
}

@article{armaniniSoftRobotsModeling2023,
  title = {Soft {{Robots Modeling}}: {{A Structured Overview}}},
  shorttitle = {Soft {{Robots Modeling}}},
  author = {Armanini, Costanza and Boyer, Fr{\'e}d{\'e}ric and Mathew, Anup Teejo and Duriez, Christian and Renda, Federico},
  year = 2023,
  month = jun,
  journal = {IEEE Transactions on Robotics},
  volume = {39},
  number = {3},
  pages = {1728--1748},
  issn = {1941-0468},
  doi = {10.1109/TRO.2022.3231360}
}

@article{chenExplorationSwimmingPerformance2020,
  title = {Exploration of Swimming Performance for a Biomimetic Multi-Joint Robotic Fish with a Compliant Passive Joint},
  author = {Chen, Di and Wu, Zhengxing and Dong, Huijie and Tan, Min and Yu, Junzhi},
  year = 2020,
  month = dec,
  journal = {Bioinspiration \& Biomimetics},
  volume = {16},
  number = {2},
  pages = {026007},
  publisher = {IOP Publishing},
  issn = {1748-3190},
  doi = {10.1088/1748-3190/abc494},
  langid = {english}
}

@article{khatibOceanOneRobotic2016,
  title = {Ocean {{One}}: {{A Robotic Avatar}} for {{Oceanic Discovery}}},
  shorttitle = {Ocean {{One}}},
  author = {Khatib, Oussama and Yeh, Xiyang and Brantner, Gerald and Soe, Brian and Kim, Boyeon and Ganguly, Shameek and Stuart, Hannah and Wang, Shiquan and Cutkosky, Mark and Edsinger, Aaron and Mullins, Phillip and Barham, Mitchell and Voolstra, Christian R. and Salama, Khaled Nabil and L'Hour, Michel and Creuze, Vincent},
  year = 2016,
  month = dec,
  journal = {IEEE Robotics \& Automation Magazine},
  volume = {23},
  number = {4},
  pages = {20--29},
  issn = {1558-223X},
  doi = {10.1109/MRA.2016.2613281}
}

@article{laschiModelingEmbodiedIntelligence2023,
  title = {Modeling Embodied Intelligence: Can We Capture Its Essence by Modeling Internal and External Interactions?},
  shorttitle = {Modeling Embodied Intelligence},
  author = {Laschi, Cecilia},
  year = 2023,
  month = oct,
  journal = {IOP Conference Series: Materials Science and Engineering},
  volume = {1292},
  number = {1},
  pages = {012001},
  publisher = {IOP Publishing},
  issn = {1757-899X},
  doi = {10.1088/1757-899X/1292/1/012001},
  langid = {english}
}

@article{liaoDynamicModelingPerformance2022,
  title = {Dynamic {{Modeling}} and {{Performance Analysis}} for a {{Wire-Driven Elastic Robotic Fish}}},
  author = {Liao, Xiaocun and Zhou, Chao and Zou, Qianqian and Wang, Jian and Lu, Ben},
  year = 2022,
  month = oct,
  journal = {IEEE Robotics and Automation Letters},
  volume = {7},
  number = {4},
  pages = {11174--11181},
  issn = {2377-3766},
  doi = {10.1109/LRA.2022.3197911}
}

@article{liuUnderwaterSoftArm2026,
  title = {Underwater Soft Arm Grasping with Simplified Control Using Octopus-Inspired Bending Propagation},
  author = {Liu, Jiaqi and Zhu, Zhichao and Wen, Li},
  year = 2026,
  month = jan,
  journal = {npj Robotics},
  volume = {4},
  number = {1},
  pages = {2},
  publisher = {Nature Publishing Group},
  issn = {2731-4278},
  doi = {10.1038/s44182-025-00066-9},
  copyright = {2025 The Author(s)},
  langid = {english}
}

@article{mathewReducedOrderModeling2025,
  title = {Reduced Order Modeling of Hybrid Soft-Rigid Robots Using Global, Local, and State-Dependent Strain Parameterization},
  author = {Mathew, Anup Teejo and {Feliu-Talegon}, Daniel and Alkayas, Abdulaziz Y and Boyer, Frederic and Renda, Federico},
  year = 2025,
  month = jan,
  journal = {The International Journal of Robotics Research},
  volume = {44},
  number = {1},
  pages = {129--154},
  publisher = {SAGE Publications Ltd STM},
  issn = {0278-3649},
  doi = {10.1177/02783649241262333},
  langid = {english}
}

@article{mazzeoMarineRoboticsDeepSea2022,
  title = {Marine {{Robotics}} for {{Deep-Sea Specimen Collection}}: {{A Systematic Review}} of {{Underwater Grippers}}},
  shorttitle = {Marine {{Robotics}} for {{Deep-Sea Specimen Collection}}},
  author = {Mazzeo, Angela and Aguzzi, Jacopo and Calisti, Marcello and Canese, Simonepietro and Vecchi, Fabrizio and Stefanni, Sergio and Controzzi, Marco},
  year = 2022,
  month = jan,
  journal = {Sensors},
  volume = {22},
  number = {2},
  pages = {648},
  publisher = {Multidisciplinary Digital Publishing Institute},
  issn = {1424-8220},
  doi = {10.3390/s22020648},
  copyright = {http://creativecommons.org/licenses/by/3.0/},
  langid = {english}
}

@article{morganAutonomousUnderwaterManipulation2022,
  title = {Autonomous {{Underwater Manipulation}}: {{Current Trends}} in {{Dynamics}}, {{Control}}, {{Planning}}, {{Perception}}, and {{Future Directions}}},
  shorttitle = {Autonomous {{Underwater Manipulation}}},
  author = {Morgan, Edward and Carlucho, Ignacio and Ard, William and Barbalata, Corina},
  year = 2022,
  month = dec,
  journal = {Current Robotics Reports},
  volume = {3},
  number = {4},
  pages = {187--198},
  issn = {2662-4087},
  doi = {10.1007/s43154-022-00089-2},
  langid = {english}
}

@article{ramirez-llodraDeepDiverseDefinitely2010,
  title = {Deep, Diverse and Definitely Different: Unique Attributes of the World's Largest Ecosystem},
  shorttitle = {Deep, Diverse and Definitely Different},
  author = {{Ramirez-Llodra}, E. and Brandt, A. and Danovaro, R. and De Mol, B. and Escobar, E. and German, C. R. and Levin, L. A. and Martinez Arbizu, P. and Menot, L. and {Buhl-Mortensen}, P. and Narayanaswamy, B. E. and Smith, C. R. and Tittensor, D. P. and Tyler, P. A. and Vanreusel, A. and Vecchione, M.},
  year = 2010,
  month = sep,
  journal = {Biogeosciences},
  volume = {7},
  number = {9},
  pages = {2851--2899},
  publisher = {Copernicus GmbH},
  issn = {1726-4170},
  doi = {10.5194/bg-7-2851-2010},
  langid = {english}
}

@inproceedings{todorovMuJoCoPhysicsEngine2012,
  title = {{{MuJoCo}}: {{A}} Physics Engine for Model-Based Control},
  shorttitle = {{{MuJoCo}}},
  booktitle = {2012 {{IEEE}}/{{RSJ International Conference}} on {{Intelligent Robots}} and {{Systems}}},
  author = {Todorov, Emanuel and Erez, Tom and Tassa, Yuval},
  year = 2012,
  month = oct,
  pages = {5026--5033},
  issn = {2153-0866},
  doi = {10.1109/IROS.2012.6386109}
}

@article{wangShallowWatersMariana2026,
  title = {From Shallow Waters to {{Mariana Trench}}: {{A}} Survey of Bio-Inspired Underwater Soft Robots},
  shorttitle = {From Shallow Waters to {{Mariana Trench}}},
  author = {Wang, Jie and Du, Peng and Zhang, Yiyuan and Xie, Zhexin and Laschi, Cecilia},
  year = 2026,
  journal = {Bioinspiration \& Biomimetics},
  issn = {1748-3190},
  doi = {10.1088/1748-3190/ae3af2},
  langid = {english}
}

@article{wangSpiRobsLogarithmicSpiralshaped2025,
  title = {{{SpiRobs}}: {{Logarithmic}} Spiral-Shaped Robots for Versatile Grasping across Scales},
  shorttitle = {{{SpiRobs}}},
  author = {Wang, Zhanchi and Freris, Nikolaos M. and Wei, Xi},
  year = 2025,
  month = apr,
  journal = {Device},
  volume = {3},
  number = {4},
  publisher = {Elsevier},
  issn = {2666-9994, 2666-9986},
  doi = {10.1016/j.device.2024.100646},
  langid = {english}
}

@misc{wang_openspirob,
  title        = {OpenSpiRobs: Open Spiral Robots Toolkit},
  author       = {Wang, Zhanchi},
  year         = {2026},
  url          = {https://github.com/ZhanchiWang/Open-Spiral-Robots}
}

@inproceedings{zhangOctopusSwimmingLikeRobotSoft2025a,
  title = {Octopus-{{Swimming-Like Robot}} with {{Soft Asymmetric Arms}}},
  booktitle = {2025 {{IEEE}} 8th {{International Conference}} on {{Soft Robotics}} ({{RoboSoft}})},
  author = {Zhang, Bobing and Zhang, Yiyuan and Li, Yiming and Xuan, Sicheng and Ng, Hong Wei and Liufu, Yuliang and Tang, Zhiqiang and Laschi, Cecilia},
  year = 2025,
  month = apr,
  pages = {1--8},
  issn = {2769-4534},
  doi = {10.1109/RoboSoft63089.2025.11020919}
}

@article{zhongDesignModelingExperiment2024,
  title = {Design, {{Modeling}}, and {{Experiment}} of {{Underactuated Flexible Gliding Robotic Fish}}},
  author = {Zhong, Yong and Wang, Qixin and Yang, Jiawei and Wang, Chengcai},
  year = 2024,
  month = jun,
  journal = {IEEE/ASME Transactions on Mechatronics},
  volume = {29},
  number = {3},
  pages = {2266--2276},
  issn = {1941-014X},
  doi = {10.1109/TMECH.2023.3328034}
}

@inproceedings{
sobanbabu2025samplingbased,
title={Sampling-based System Identification with Active Exploration for Legged Sim2Real Learning},
author={Nikhil Sobanbabu and Guanqi He and Tairan He and Yuxiang Yang and Guanya Shi},
booktitle={9th Annual Conference on Robot Learning},
year={2025},
url={https://openreview.net/forum?id=UTPBM4dEUS}
}

@misc{ravi2024sam,
  title={Sam 2: Segment anything in images and videos},
  author={Ravi, Nikhila and Gabeur, Valentin and Hu, Yuan-Ting and Hu, Ronghang and Ryali, Chaitanya and Ma, Tengyu and Khedr, Haitham and R{\"a}dle, Roman and Rolland, Chloe and Gustafson, Laura and others},
  journal={arXiv preprint arXiv:2408.00714},
  year={2024}
}

@article{li_bioinspired_2026,
    title = {Bioinspired underwater soft robots: from biology to robotics and back},
    volume = {4},
    issn = {2731-4278},
    shorttitle = {Bioinspired underwater soft robots},
    number = {1},
    journal = {npj Robotics},
    author = {Li, Lei and Qin, Boyang and Gao, Wenzhuo and Li, Yanyu and Zhang, Yiyuan and Wang, Bo and Kong, Shihan and Wang, Jian and He, Dekui and Yu, Junzhi},
    month = apr,
    year = {2026},
    pages = {25},
}

@inproceedings{huang_physical_2026,
    title = {Physical {Human}-{Robot} {Interaction} for {Grasping} in {Augmented} {Reality} via {Rigid}-{Soft} {Robot} {Synergy}},
    booktitle = {2026 {IEEE} 9th {International} {Conference} on {Soft} {Robotics} ({RoboSoft})},
    publisher = {IEEE},
    author = {Huang, Huishi and Klusmann, Jack and Wang, Haozhe and Ji, Shuchen and Ying, Fengkang and Zhang, Yiyuan and Nassour, John and Cheng, Gordon and Rus, Daniela and Liu, Jun and Ang, Marcelo H. and Laschi, Cecilia},
    month = apr,
    year = {2026},
    pages = {1468--1475},
}

\end{document}